\newif\ifieee
\newif\ifrecall
\newif\iffastnn
\newif\ifbrief

\fastnntrue
\recalltrue
\ieeefalse
\brieftrue

\ifieee
\documentclass[journal]{IEEEtran}
\else
\documentclass[runningheads]{llncs}
\fi

\usepackage{graphicx}
\graphicspath{{./figures/}{./figures/plots/}}
\usepackage[caption=false,font=footnotesize]{subfig}

\usepackage{hyperref}
\usepackage{amsmath}
\usepackage{url}

\ifieee
\usepackage{array}
\usepackage[caption=false,font=footnotesize]{subfig}
\usepackage{dblfloatfix}
\fi

\usepackage{todonotes}
\usepackage{soul}

\newcommand{\HF}{HF}

\newcommand{\oneshot}{one-shot}
\newcommand{\Oneshot}{One-shot}
\newcommand{\oneshotGen}{\oneshot\ classification}
\newcommand{\OneshotGen}{\Oneshot\ classification}
\newcommand{\oneshotInst}{\oneshot\ instance-classification}
\newcommand{\OneshotInst}{\Oneshot\ instance-classification}

\newcommand{\PC}{PC}
\newcommand{\PS}{PS}
\newcommand{\PR}{PR}
\newcommand{\VC}{LTM}
\newcommand{\PM}{PM}
\newcommand{\FNN}{FastNN}

\begin{document}

\title{Unsupervised One-shot Learning of Both Specific Instances and Generalised Classes with a Hippocampal Architecture}

\ifieee
\author{Gideon~Kowadlo,
        Abdelrahman~Ahmed,
        and~David~Rawlinson
\thanks{Gideon Kowadlo, Abdelrahman Ahmed and David Rawlinson were with ProjectAGI, research lab of Incubator491 Melbourne, Australia. web: https://agi.io, email: gideon/abdel/dave@agi.io}
\thanks{Manuscript submitted July, 2020}}

\else

\titlerunning{One-shot Learning of Instances and Generalised Classes with AHA}
\author{Gideon Kowadlo\inst{1}\orcidID{0000-0001-6036-1180} \and
Abdelrahman Ahmed\inst{2,3}\orcidID{0000-0001-9530-766X} \and
David Rawlinson\inst{3}\orcidID{0000-0001-9443-3840}}
\authorrunning{G. Kowadlo et al.} 

\institute{Cerenaut, Melbourne, Australia\\
\email{info@cerenaut.ai}\\
\url{https://cerenaut.ai}}
\fi

\hyphenation{auto-associative}

\maketitle              

\begin{abstract}

Established experimental procedures for one-shot machine learning do not test the ability to learn or remember specific instances of classes, a key feature of animal intelligence. Distinguishing specific instances is necessary for many real-world tasks, such as remembering which cup belongs to you. Generalisation within classes conflicts with the ability to separate instances of classes, making it difficult to achieve both capabilities within a single architecture. We propose an extension to the standard Omniglot classification-generalisation framework that additionally tests the ability to distinguish specific instances after one exposure and introduces noise and occlusion corruption. Learning is defined as an ability to classify as well as recall training samples. Complementary Learning Systems (CLS) is a popular model of mammalian brain regions believed to play a crucial role in learning from a single exposure to a stimulus. We created an artificial neural network implementation of CLS and applied it to the extended Omniglot benchmark. Our unsupervised model demonstrates comparable performance to existing supervised ANNs on the Omniglot classification task (requiring generalisation), without the need for domain-specific inductive biases. On the extended Omniglot instance-recognition task, the same model also demonstrates significantly better performance than a baseline nearest-neighbour approach, given partial occlusion and noise.

\ifieee
\else
\keywords{CLS \and Hippocampus \and One-shot \and Specifics \and Instances \and Unsupervised \and Generalisation.}
\fi

\end{abstract}

\ifieee
\begin{IEEEkeywords}
CLS, Hippocampus, One-shot, Specifics, Instances, Unsupervised.
\end{IEEEkeywords}

%
\IEEEpeerreviewmaketitle

\section{Impact Statement}

This research is a first step in the development of \oneshot\ learning that includes the ability to identify specific instances. A capability that underpins the recognition of objects, things, facts as well as a personal history and with it an identity. Such a capability is a crucial component for AI to enable autonomous agents to interact with humans in complex dynamic environments, including industry as well as in people's homes and personal lives.

In addition, this research constitutes the first use of CLS (the standard model for neocortical-hippocampal interaction) on a realistic dataset. It furthers confidence in CLS and sets the ground for its use in consolidation of memories without catastrophic forgetting for Continual Learning, another major challenge to the adoption of ML for meaningful AI.

\fi

\section{Introduction}

\Oneshot\ learning has seen renewed interest in recent years.
Many studies \cite{Lake2015,Vinyals2016} are motivated by the apparent limitations of modern ML relative to animal-like learning \cite{Lake2017}.
An ability for \oneshot\ learning alleviates the reliance on large labelled datasets in which samples are assumed to be i.i.d., implying an unchanging world.
This is particularly relevant for autonomous real-world agents, where samples are necessarily highly correlated and typically unlabelled.

However, we believe that the standard approach - classification of general classes - does not go far enough.
Learning specific instances is crucial for intelligent agents, and is something we take for granted.
For example, identifying your own coffee cup from other cups, in addition to recognising that it belongs to the `cup' category. More generally, it underpins memory for singular facts and an individual's own autobiographical history, important for future decision making.


At first glance, learning specific instances appears easy. An obvious starting point is nearest-neighbour lookup in a buffer of past observations.
However, this approach may perform poorly given observational variation such as occlusion, and have trouble also generalising class recognition ability.
Conversely, methods that can generalise would be unlikely to do well at learning specific instances, as they are conflicting capabilities.

Learning of specific instances is not to be confused with Instance-based Learning \cite{Russell2009}.
Such approaches store instances during training, and use them to classify test samples e.g. k-nearest neighbour, SVM's and RBFs.
Our objective is learning a model of the instances, despite observational variation, while being able to distinguish even very similar instances from each other.
We identify two important aspects of learning - classification and recall (generation) of concept.

Complementary Learning Systems (CLS) is a model of mammalian learning that describes the interplay between the neocortex and a region called the Hippocampal Formation (\HF) \cite{McClelland1995,Rolls1995}. CLS is believed to be crucial for fast learning and is recognised to be important for intelligent agents \cite{Kumaran2016}.
Our motivation is to expand the standard definition of \oneshot\ learning, and to test if the CLS architecture can satisfy the requirements.

In this paper we propose a broader benchmark for \oneshot\ learning that includes robust classification of specific instances given observational variance  by introducing image corruption with occlusion and noise. 
We present an ANN implementation of CLS using an Artificial Hippocampal Algorithm (AHA) and apply it to the extended benchmark.
The performance of the system is compared to two baselines, a simplified version of CLS that replaces the hippocampal model with a conventional ML model optimised for the task, and to the naive solution for learning specifics - a buffer with nearest neighbour lookup. 


\section{Background}
\label{sec:background}

\subsection{\Oneshot\ Learning}
\label{sec:background_oneshot}
Following seminal work by Li et al. \cite{Fei-Fei2003,Fei-Fei2006}, the area was re-invigorated by Lake et al. \cite{Lake2011}, who introduced a popular test that has become a standard benchmark \cite{Lake2015}. It is a \oneshot\ classification task on the Omniglot dataset of handwritten characters. Classification is posed as a matching task, where a given character must be matched with a character of the same class in a test set, see Section~\ref{sec:experimental_method} for details. The framework was formalised by Vinyals et al. \cite{Vinyals2016}.

A typical approach is to pre-train a model on many classes and use learnt concepts to recognise new classes quickly from one or few examples. Often framed as meta-learning or ``learning to learn'', there are multiple implementations using neural networks that require external labels and supervised learning during pre-training, such as Siamese networks \cite{Koch2015a}, matching networks \cite{Vinyals2016}, and prototypical networks \cite{Snell2017}. 
Two notable Bayesian approaches, BPL \cite{Lake2015} and RCN \cite{George2017}, achieve above and close to human level performance respectively. 
The superior performance of BPL may be partially explained by its use of prior knowledge about handwriting via stroke formation. RCN, by virtue of the design which is modeled on the visual cortex, is also specialised for this type of visual task. It is less clear how it could be applied to other datasets and problems where contour topology is less distinct or relevant.
A comprehensive review is given in \cite{Lake2019}.

\subsection{Complementary Learning Systems (CLS)}
\label{sec:background_cls}
Complementary Learning Systems (CLS) is a standard framework for understanding the function of the \HF\ \cite{McClelland1995,OReilly2014,Kumaran2016}. CLS consists of two differentially specialised and complementary structures, neocortex and \HF, shown in Figure~\ref{fig:complementary_systems}.
In this framework, the neocortex is analogous to a conventional ML model, incrementally learning regularities across many observations, comprising a long-term memory (LTM). It forms overlapping and distributed representations that are effective for inference. In contrast, the \HF\ rapidly learns distinct observations, forming sparser, non-overlapping and therefore non-interfering representations, functioning as a short term memory (STM). Recent memories from the \HF\ are replayed to neocortex, re-instating the original activations resulting in consolidation as long-term memory (LTM). Patterns are replayed in an interleaved fashion, avoiding catastrophic forgetting. In addition, they can be replayed selectively according to salience. 
There have been numerous implementations of CLS \cite{Norman2003,Ketz2013,Greene2013,Rolls2013,Schapiro2017a} and Rolls et al. presented a similar model with greater neuroanatomical detail \cite{Rolls1995}.

Overall, the \HF\ functions as an autoassociative memory that can recall memories from partial cues.
CLS describes the \HF\ in terms of distinct functional units called subfields. 
Together they comprise a unification of pattern completion and pattern separation pathways. 
Reported implementations (see citations above) are expressed at the level of individual neurons replicating known biological plasticity and dynamics, and have not been applied to ML benchmarks.

\section{Model}
\label{sec:methodology}

Our approach is to implement a CLS-style STM with an LTM (Figure~\ref{fig:complementary_systems}), with biological plausibility constraints \cite{Kowadlo2019} - all components are trained with local and immediate credit assignment.
The LTM comprises a simple vision component suitable for image feature extraction.
The STM is implemented with AHA, an Artificial Hippocampal Algorithm, which follows the subfield architecture of the \HF\ described by CLS.
There are two baselines for comparison.
Firstly, the LTM alone is compared to performance of LTM+STM.
It constitutes a naive solution to classifying specific instances (see Section~\ref{sec:experimental_method}). 
The second baseline, \FNN, is an alternate STM comprised of a standard ML component empirically optimised for the same tasks. 
We provide the code and configuration required to reproduce experiments in a GitHub repository\footnote{\url{https://github.com/Cerenaut/aha}}.

\ifieee
Our hypothesis is that unsupervised LTM+AHA is more effective at the full set of tasks (Section~\ref{sec:experimental_method}) than state of-the-art approaches (Section~\ref{sec:background_oneshot}), which are suited to general classes.
Secondly, that a CLS-style STM will be more effective than the alternate \FNN\ and that both complementary architectures of an LTM+STM will have advantages over the LTM alone.
Secondly, that a CLS-style STM will have advantages over the LTM alone, which is suited to specific instances.
\fi


\subsection{Training and Testing Framework}
\label{sec:train_test_fwk}

With complementary systems, training and testing are non-standard, and are therefore explained here to provide context for the remainder of the paper.

\noindent \textbf{Stage 1: Pre-train LTM}: Train LTM on a training set over multiple epochs.
The LTM learns incrementally about common features that can be used compositionally to represent unseen classes.

\noindent \textbf{Stage 2: Evaluate LTM+STM}: Evaluation is conducted with a disjoint evaluation set. The LTM does not learn during this stage. Training and Testing of STM occurs rapidly, allowing multiple internal cycles but only one exposure to an external stimulus. The STM is reset after each evaluation\footnote{Adam Optimizer is reset and trainable parameters are re-initialised.}. Evaluation consists of two steps performed in succession - Train (encoding) and Test (inference).  

\begin{itemize}
	\item \textit{Train}: A small support set is presented once (referred to as `study set' in CLS). STM modules are set to train mode to learn the samples. 
	\item \textit{Test}: A small query set is presented (referred to as `recall' in CLS), STM modules are in inference mode. For each `recall' sample, the system is expected to retrieve the corresponding sample from the `study' set. If correct, it is considered to be `recognised' - an AHA moment!
\end{itemize}

\subsection{LTM - Vision Component}
\label{sec:vc}
The role of the LTM is to process high-dimensional sensor input, and output relatively abstract visual features that can be used as compositional primitives.
A single layer convolutional sparse autoencoder based on \cite{Makhzani2013,Makhzani2015} provides the required embedding.
However, in Omniglot there is a lot of empty background that is encoded with strong hidden layer activity. Lacking an attention mechanism, this detracts from compositionality of foreground features. To suppress encoding of the background, we added an `Interest Filter' which loosely mimics known retinal processing (see below). Smoothing is applied to provide some tolerance to feature location and a final max-pooling stage to reduce dimensionality. 

\paragraph{Interest Filter} The retina possesses centre-surround inhibitory and excitatory cells that can be well approximated with a Difference of Gaussians (DoG) kernel \cite{Young1987}. 
Positive and negative DoG filters are used to enhance positive and negative intensity transitions. Local non-maxima suppression merges nearby features and a `top-$k$' function creates a mask of the most significant features globally.
Positive and negative masks are combined by summation giving a final 2D mask that is applied to all channels of the convolutional autoencoder output.

\subsection{STM - Artificial Hippocampal Algorithm (AHA)}

AHA is our implementation of CLS. For greater details on CLS, the biological basis for AHA design choices, and in-depth implementation details, see \cite{Kowadlo2019}.
The components and connectivity are shown in Figure~\ref{fig:aha}.
LTM outputs sparse distributed overlapping patterns. 
The signal becomes sparser and more orthogonal through \PS, minimising interference between patterns, resulting in distinct representations for similar inputs.

In train mode, \PS\ patterns are encoded/memorised into \PC, an autoassociative, content-addressable memory. They form a target, which \PR\ learns to retrieve from \VC\ distributed representations. \PM\ learns to map from the stored non-interfering patterns, to the originating sparse distributed patterns.

In test mode, \PR\ retrieves the corresponding stored \PC\ pattern using input from \VC, which is used to cue complete recall from \PC. \PS\ is not used as a cue, because even small input differences will result in orthogonal \PS\ outputs. \PC\ can retrieve a crisp, complete pattern, that in turn enables \PM\ to recall the original observation. In future work this will be used for improved inference and consolidation of memories. 

The use of \PS\ for encoding and \PR\ for recall is based on the Hippocampal model by Rolls \cite{Rolls1995,Rolls2013}. The role of each subfield is detailed below:

\begin{figure*}[t!]
\centering
\subfloat[]{\includegraphics[width=0.3\textwidth]{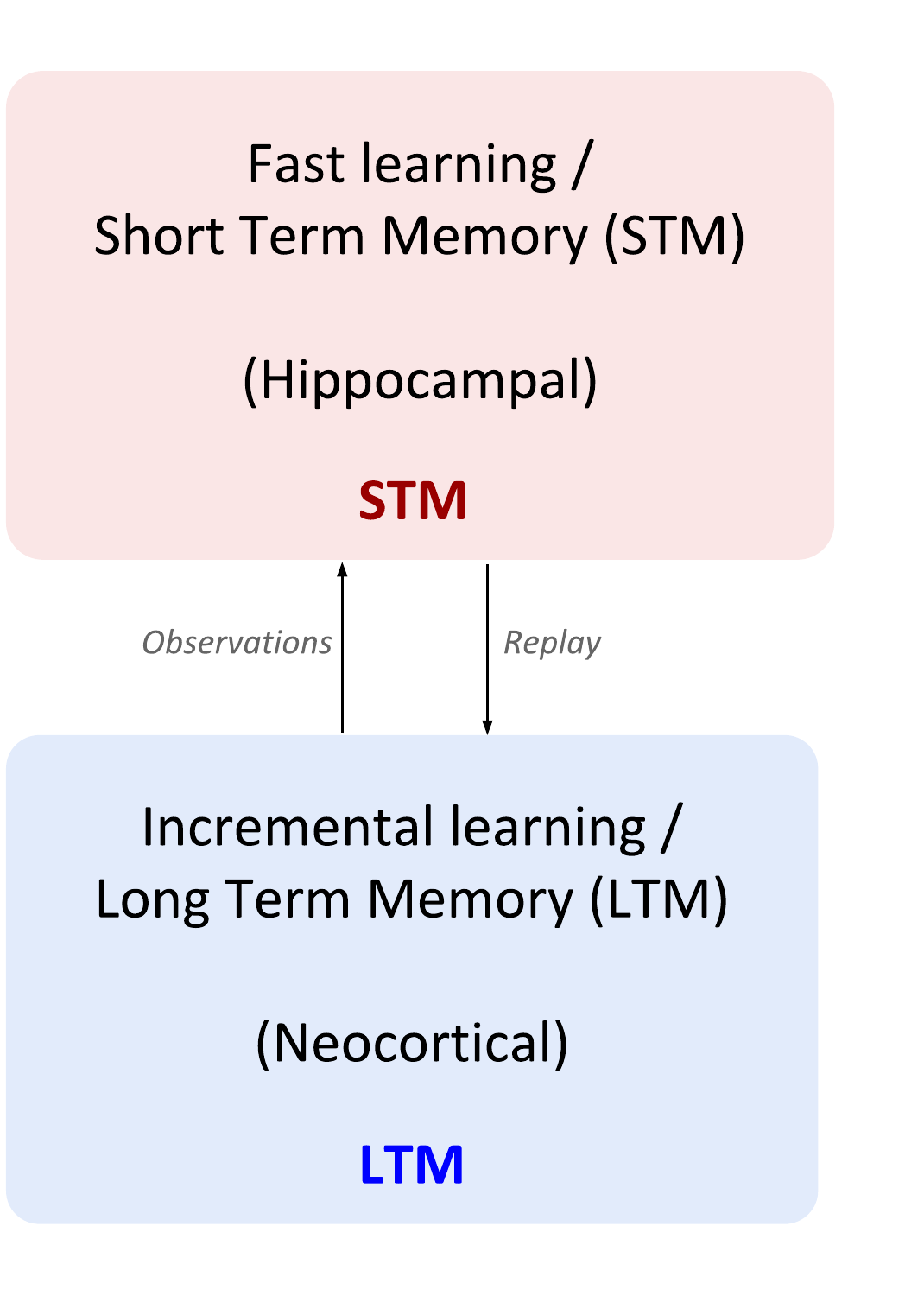}%
\label{fig:complementary_systems}}
\hfil
\subfloat[]{\includegraphics[width=0.65\textwidth]{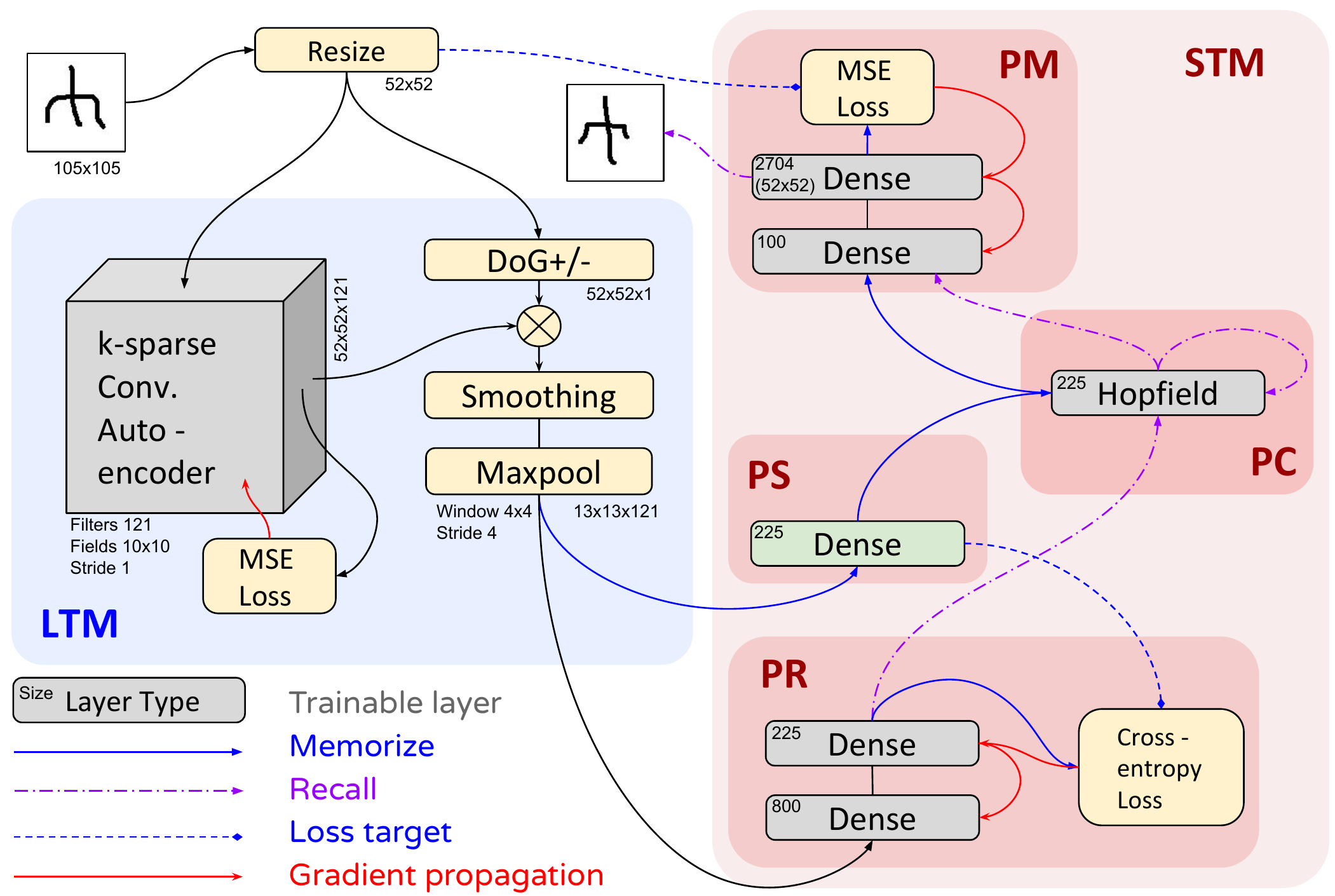}%
\label{fig:aha}}
\caption{a) \textbf{CLS}: The STM learns and forgets rapidly. Salient memories are replayed to the LTM for incremental statistical learning. b) \textbf{System diagram:} Our implementation of CLS. Local credit assignment via shallow backpropagation is used throughout. The dense layer in PS (green) is initialised, but not trained.}
\label{fig:system_diagram}
\end{figure*}

\paragraph{\PS\ - Pattern Separation}
\PS\ is implemented with a single fully-connected Artificial Neural Network (ANN) layer with sparsity constraints and temporal inhibition.
Sparsity is implemented as a `top-$k$' ranking per sample, mimicking a local competitive process via inhibitory interneurons. Low $k$ produces outputs with low overlap, but orthogonality is further improved by replicating the sparse connectivity observed in this pathway in the hippocampus \cite{Rolls2013}. A portion of the incoming connections are removed by setting weights to zero (similar to the sparsening technique of \cite{Ahmad2019}). 
Additionally, after a neuron fires (i.e. it is amongst the top-$k$), it is temporarily inhibited, mimicking the refractory period observed in biological neurons.
\PS\ is initialised with uniformly distributed random weights and does not undergo any training.

\paragraph{\PC\ - Pattern Completion}
\PC\ is implemented with a Hopfield network \cite{Hopfield1982}. 
Unlike a standard Hopfield network, there are separate input pathways for encoding (\PS) and recall (\PR).
Output layers of \PS\ and \PR\ are the same size as \PC. \PS\ and \PR\ output signals are conditioned from a continuous value [0,1] to a binary signed unit range [-1,1], chosen for better Hopfield performance.

\paragraph{\PR\ - Pattern Retrieval}
\label{sec:pr}
\PR\ is implemented with a 2-layer fully-connected ANN. 
In training, \PS\ output is used as an internally generated label constituting self-supervised learning \cite{Gidaris2019b}. 
Usually in self-supervised learning, prior task knowledge is used to set a pre-conceived goal such as rotation, with the motivation of learning generalisable representations. In the case of AHA, no prior is required. The motivation is separability and as such, the use of orthogonal patterns as labels is very effective.

\ifrecall
\paragraph{\PM\ - Pattern Mapping}
\PM\ is implemented with a 2-layer fully-connected ANN. In this study we trained it to reconstruct the input images rather than the \VC\ output, for easy assessment of recalled image quality and correctness.

\fi

\subsubsection{AHA - Theory of Operation}
\label{sec:theory_of_operation}

\paragraph{Compositionality}
\label{sec:theory_of_operation_comp}
A central capability of AHA is the memorisation of new conjunctions of primitive concepts. The primitives can be composed in a vast number of new combinations, a feature of animal-like learning \cite{Lake2017}. 
Memorisation of conjunctions of concepts is an aspect of episodic memory, as identified in the hippocampal computational modelling literature \cite{Ketz2013} (expanded in \cite{Kowadlo2019}).

Generalisation to subsequent observations of the new combination is achieved through unification of separation and completion (below).
The scope of generalisation depends on the level of abstraction of the primitives.



\paragraph{Unifying Separation and Completion}
Separation and completion are conflicting capabilities requiring separate pathways. Unification is achieved through the collaboration of \PS\ and \PR.
\PS\ sets a target for \PR\ and \PC\ to learn, providing a common representational `space'.
This makes it possible to separate \PC\ encoding and retrieval between the separation and completion pathways respectively.
In this way, they don't conflict with each other, but each operate to their strengths.



%

\subsection{Baseline STM - \FNN}
\label{sec:comparison_model}
\FNN\ is a 2-layer fully-connected ANN.
Like AHA, the target for recall is the input image itself (rather than LTM encodings) for ease of analysis.
It is `fast' in that it also learns given only one external stimulus.
We empirically optimised the learning rate, training iterations, number and size of hidden layers and the other hyperparameters.

\section{Experiments}
\label{sec:experimental_method}

\paragraph{Omniglot Benchmark - \OneshotGen\ task}
We tested our models on the \oneshotGen\ test from \cite{Lake2015}.
Referring to the train/test framework (Section~\ref{sec:train_test_fwk}), first the LTM is pre-trained on a `background' set of 30 alphabets.
Then using a disjoint `evaluation' set of 10 alphabets, a single `train' character is presented. The task is to identify the matching character from 20 distinct `test' characters from the same alphabet by a different writer. This is repeated for 20 characters comprising a single `run'. The experiment consists of 20 runs in total. Accuracy is averaged over the 20 runs.
Characters and alphabets were selected to maximise difficulty through confusion of similar characters \cite{Lake2015}. 


The method used to determine the matching character varies between reported studies. In this work, we use minimum MSE of an internal representation.
For LTM, we used the autoencoder encoding, for LTM+AHA, we report \PR\ and \PC\ and for LTM+\FNN\ the hidden layer encoding.
None of these networks are explicitly trained to classify.
In addition to accuracy, the quality of end-to-end retrieval of appropriate memories is assessed with MSE recall loss. 

\paragraph{\OneshotInst\ task}
We extended the experiments with the \oneshotInst\ task. 
It is the same as \oneshotGen, except that the `train' character exemplar must be matched with the exact same exemplar amongst 20 `test' distractor exemplars of the \emph{same} character class. 
Being the same character class, all the exemplars are very similar making separation difficult.
In each run, the character class and exemplars are selected by randomly sampling without repeats from the `evaluation' set.

\paragraph{Common Conditions}
In addition, we explored robustness by introducing image corruption to emulate realistic challenges in visual processing that could also apply to other sensory modalities.
Noise emulates imperfect sensor capture. For example, in visual recognition, the target object might be dirty or lighting conditions changed. Occlusion emulates incomplete sensing e.g. due to obstruction by another object. Robust performance is a feature of animal-like learning that would confer practical benefits to machines, and is therefore important to explore \cite{Ahmad2019}.
Occlusion is achieved with randomly placed circles, completely contained within the image. 
Noise is introduced by replacing a proportion of the pixels with a random intensity value drawn from a uniform distribution.

For both tests, instead of presenting 1 test character at a time, all 20 are presented simultaneously, made possible by the short term memory of CLS. 
Noise and occlusion is increased from none, to almost complete corruption, in 10 increments. The highest level is capped at 98\% corruption, to ensure some meaningful output. 
Every test is repeated with 10 random seeds.

In \oneshotGen, strong generalisation is required as well as some pattern separation to distinguish similar character classes.
The \oneshotInst\ task requires strong pattern separation, as well as some generalisation for robustness.

\section{Results}
\label{sec:results}

\begin{figure*}[t!]
\centering
\subfloat[\OneshotGen\ with occlusion]{\includegraphics[width=0.49\textwidth]{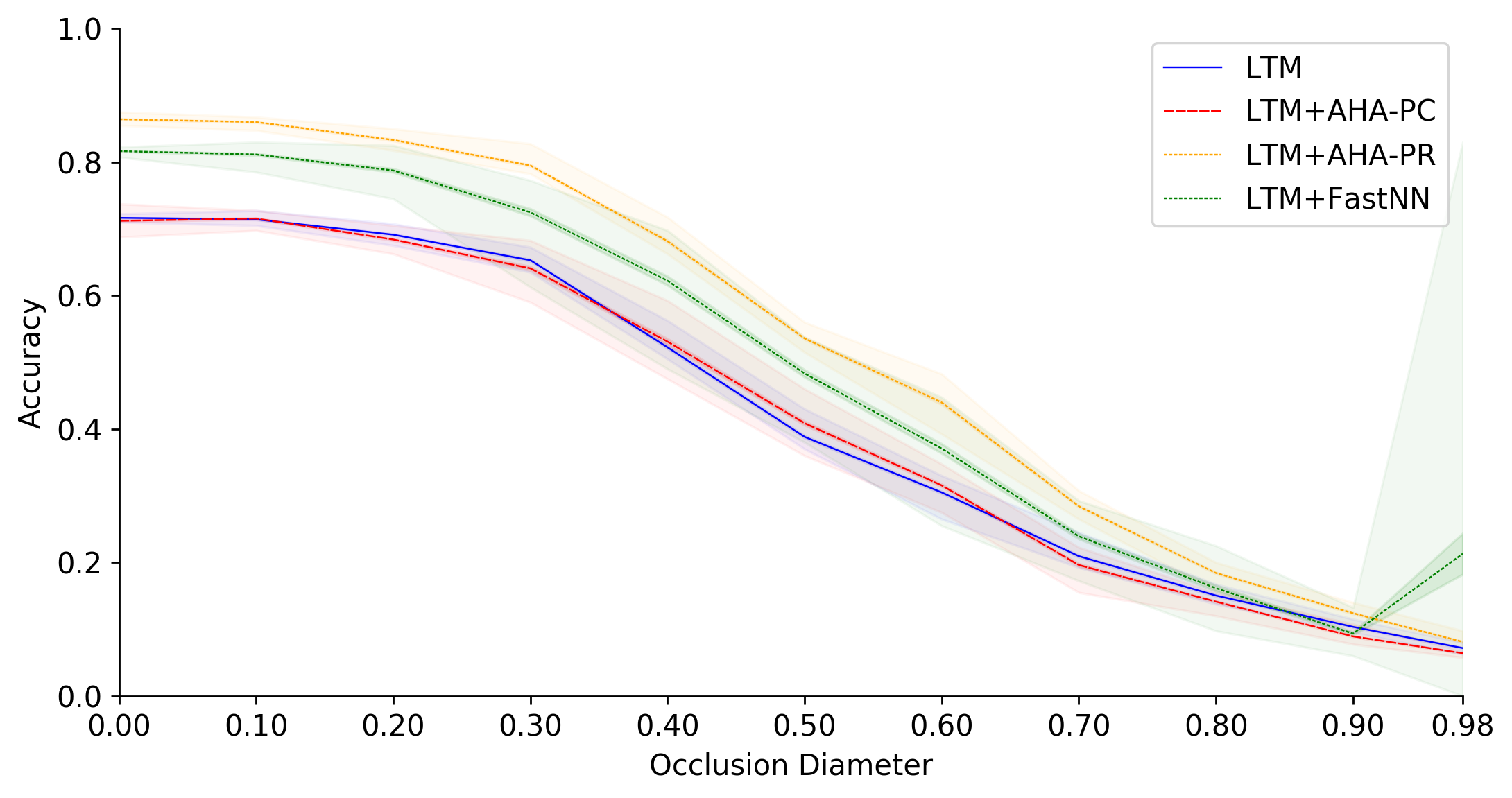}%
\label{fig:accuracy_plots_class_occ}}
\hfil
\subfloat[\OneshotGen\ with noise]{\includegraphics[width=0.49\textwidth]{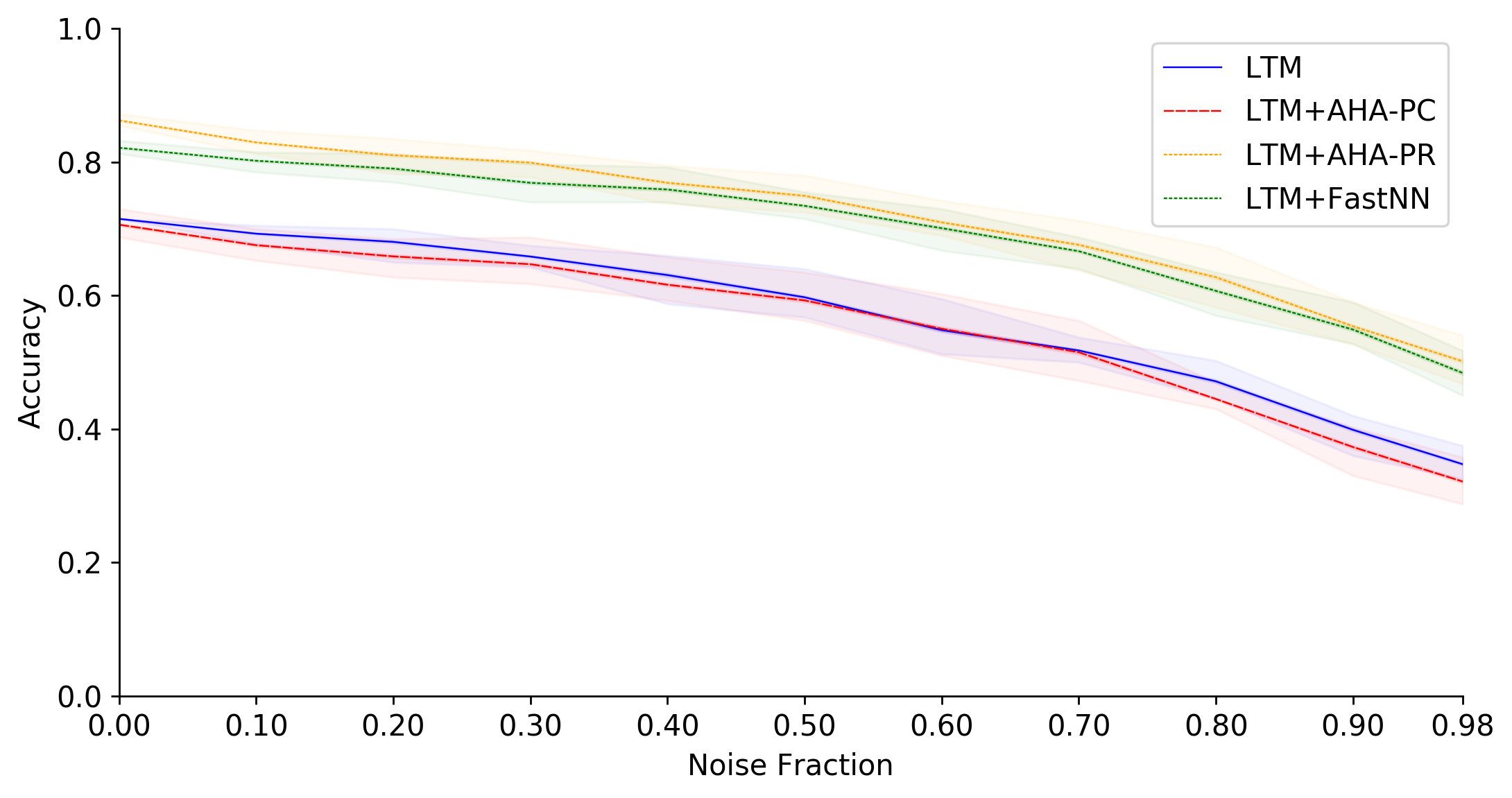}%
\label{fig:accuracy_plots_class_noise}}

\subfloat[\OneshotInst\ with occlusion]{\includegraphics[width=0.49\textwidth]{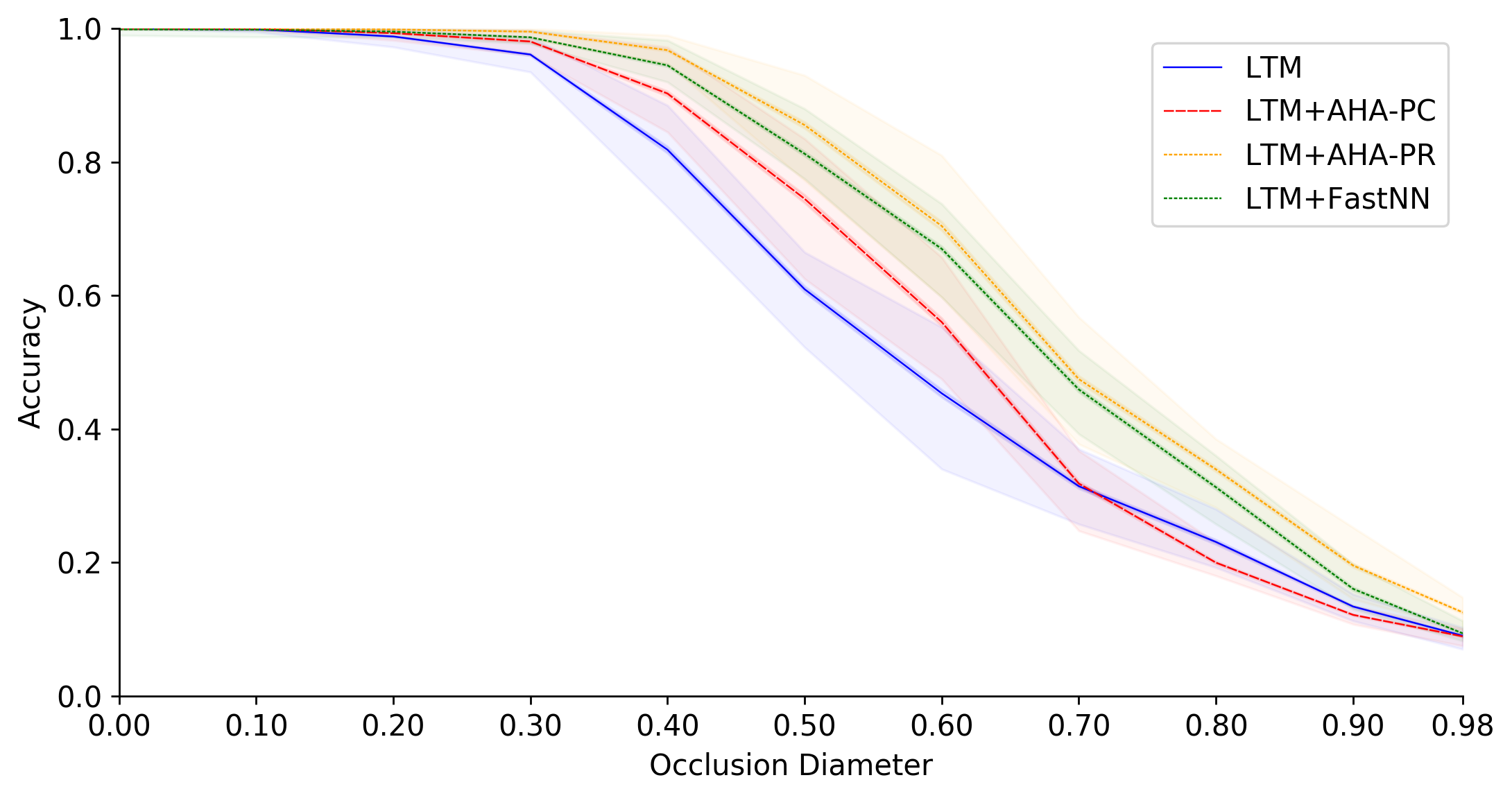}%
\label{fig:accuracy_plots_inst_occ}}
\hfil
\subfloat[\OneshotInst\ with noise]{\includegraphics[width=0.49\textwidth]{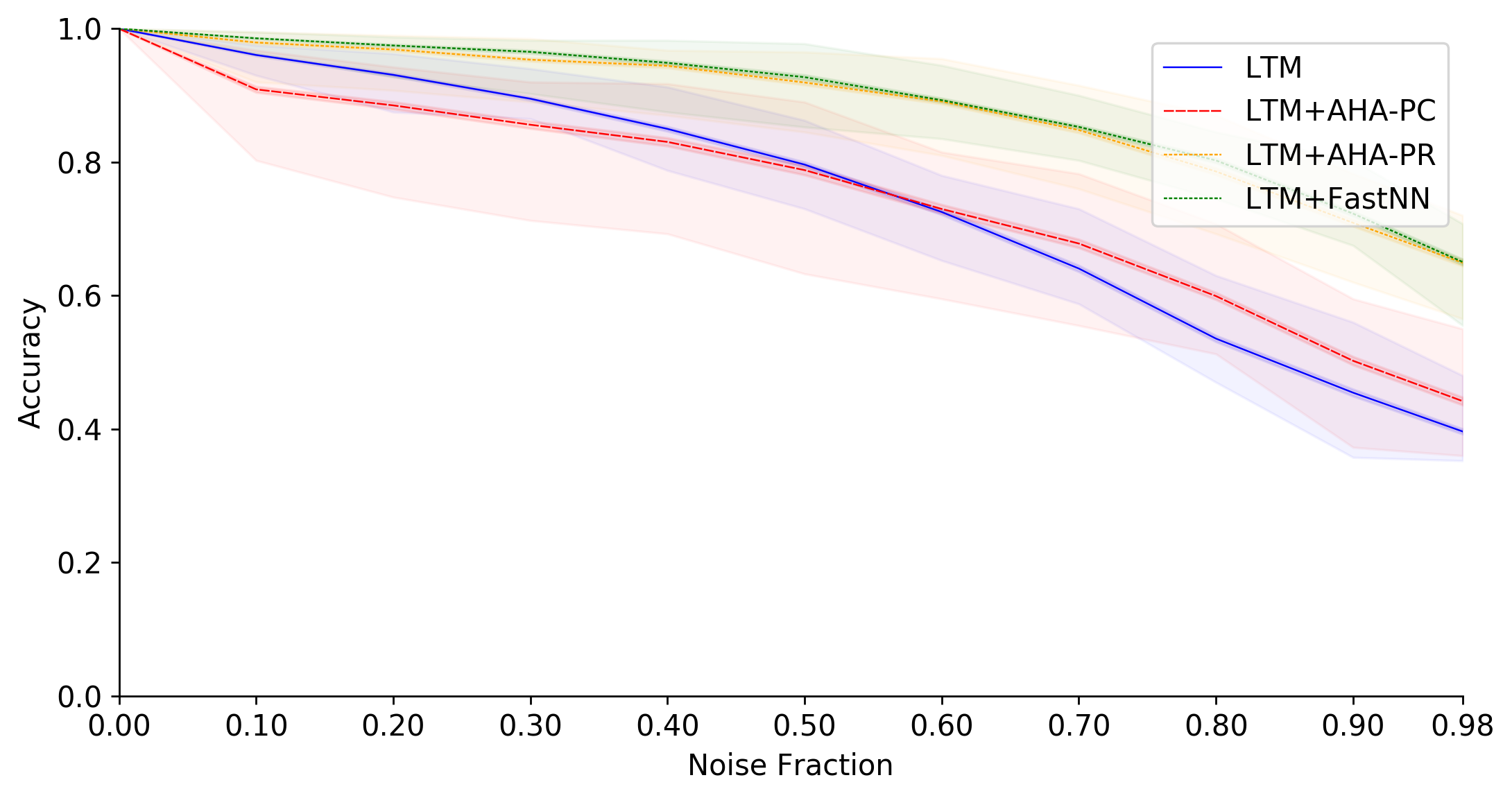}%
\label{fig:accuracy_plots_inst_noise}}
\caption{\textbf{Accuracy vs occlusion and noise.}
LTM+STM improves performance over baseline LTM. AHA STM is superior to the baseline \FNN\ STM. The effect is more pronounced for occlusion than noise.
Occlusion diameter is expressed as a fraction of image side length, noise as a fraction of image area. The mean value is bold with medium shading for 1 standard deviation, light shading demarcates min/max values.}
\label{fig:accuracy_plots}
\end{figure*}

\subsection{Accuracy}
\OneshotGen\ results are shown in Figures \ref{fig:accuracy_plots_class_occ} and \ref{fig:accuracy_plots_class_noise}.
LTM accuracy starts at $71.6\%$, without image corruption.
Increasing noise affects all features equally and gradually, whereas occlusion increases the likelihood of suddenly removing important topological features.
At very high occlusion, the character is mostly covered, leading to chance level performance.
All signals follow the same overall trend as LTM.


With no noise or occlusion, \PR\ at $86.4\%$ has an advantage of almost $15\%$ over LTM and is comparable to other ANN results for this task (Table~\ref{table:comparison}). This advantage is maintained with moderate levels of occlusion. As extreme occlusion begins to cover most of the character, the accuracy of \PR, \PC\ and LTM converge. However, the advantage is maintained over all noise levels.
\PC\ accuracy was no better than LTM.

With no corruption, \FNN\ improves on \VC\ accuracy by $10.3\%$, $4.4\%$ less than the AHA improvement. 
The advantage of AHA over \FNN\ is maintained over almost all levels of occlusion, and minor for all levels of noise.

For context, reported accuracy in the case of zero noise or occlusion is contrasted with other works in Table~\ref{table:comparison}. Existing values are reproduced from \cite{Lake2019}.

\begin{table}[t]
\begin{center}
\label{table:comparison}
\caption[Table caption text]{\textbf{Comparison of algorithms for \oneshotGen, without image corruption.} LTM+AHA is competitive with state-of-the-art convolutional approaches whilst demonstrating a wider range of capabilities.}
\begin{tabular}{cc}
\hline
\textbf{Algorithm} & \textbf{Accuracy (\%)} \\ \hline
BPL \cite{Lake2019} & 96.7 \\
\textit{Human} \cite{Lake2019} & \textit{95.5} \\
RCN \cite{George2017} & 92.7 \\
Simple Conv Net \cite{Lake2019} & 86.5 \\
\hline
\end{tabular}
\quad
\begin{tabular}{cc}
\hline
\textbf{Algorithm} & \textbf{Accuracy (\%)} \\ \hline
\textbf{LTM+AHA} & \textbf{86.4} \\
Prototypical Net \cite{Lake2019} & 86.3 \\
\textbf{LTM+\FNN} & \textbf{81.9} \\
VHE \cite{Hewitt2018} & 81.3 \\
\hline
\end{tabular}
\end{center}
\end{table}

\OneshotInst\ results are shown in Figures \ref{fig:accuracy_plots_inst_occ} and \ref{fig:accuracy_plots_inst_noise}.
LTM accuracy is perfect at low levels of image corruption, remaining almost perfect in the case of occlusion, until approximately one third of the image is affected. 
All signals follow the same trends observed for the \oneshotGen\ task.

For AHA, \PR\ accuracy remains extremely high, close to $100\%$ until a $10\%$ greater level of occlusion than for LTM i.e. addition of AHA increases tolerance to occlusion.
The advantage over LTM increases with increasing corruption, fading away for occlusion but continuing to grow for noise.

\FNN\ also improves on the baseline. It has worse accuracy than AHA for a given level of occlusion (less substantial than \oneshotGen), and almost equal accuracy for varying levels of noise.

\subsection{Recall}
\begin{figure*}[t!]
     \centering
     \subfloat[\OneshotGen\ with occlusion]{\includegraphics[width=0.49\textwidth]{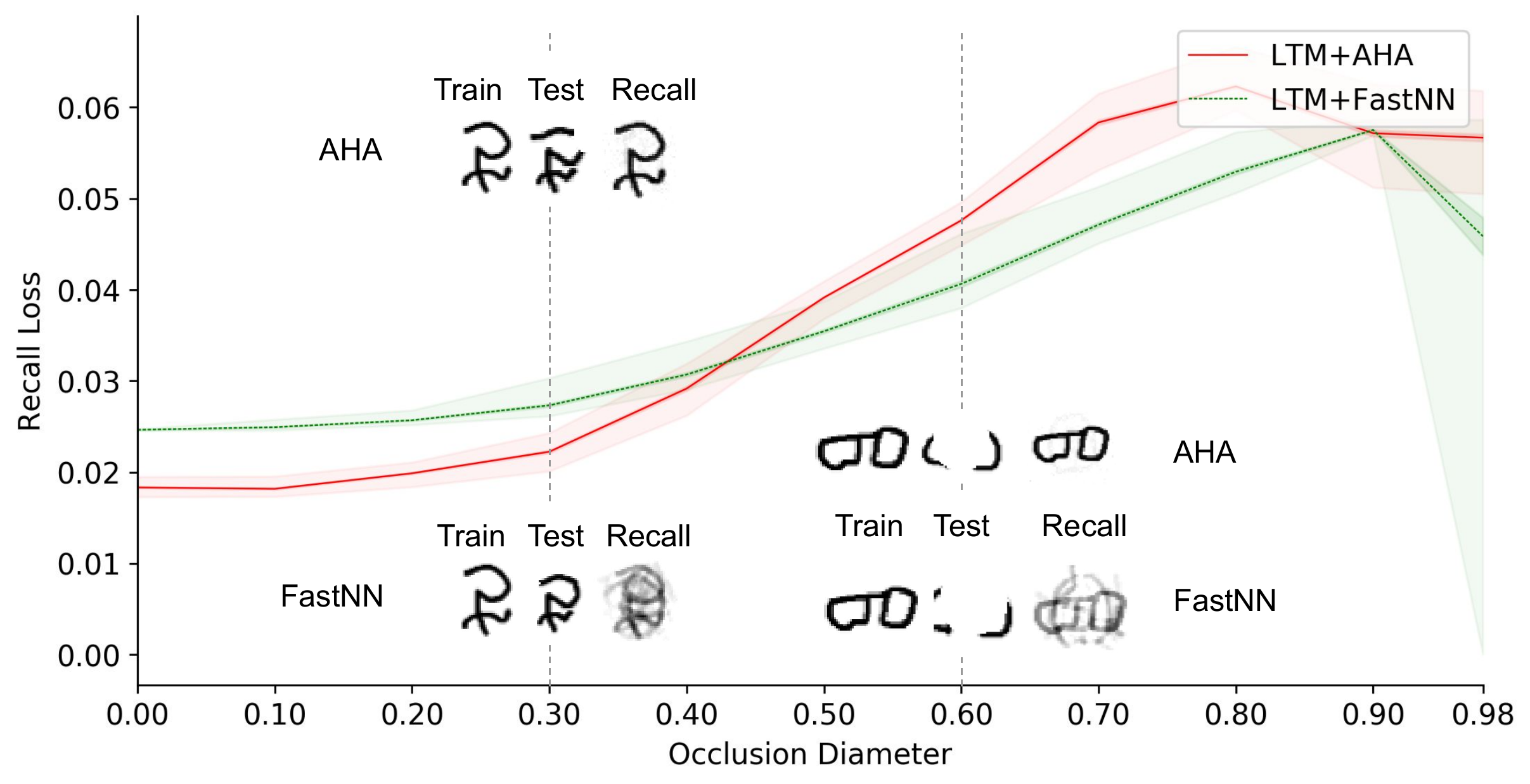}}
     \hfil
     \subfloat[\OneshotGen\ with noise]{
		 \includegraphics[width=0.49\textwidth]{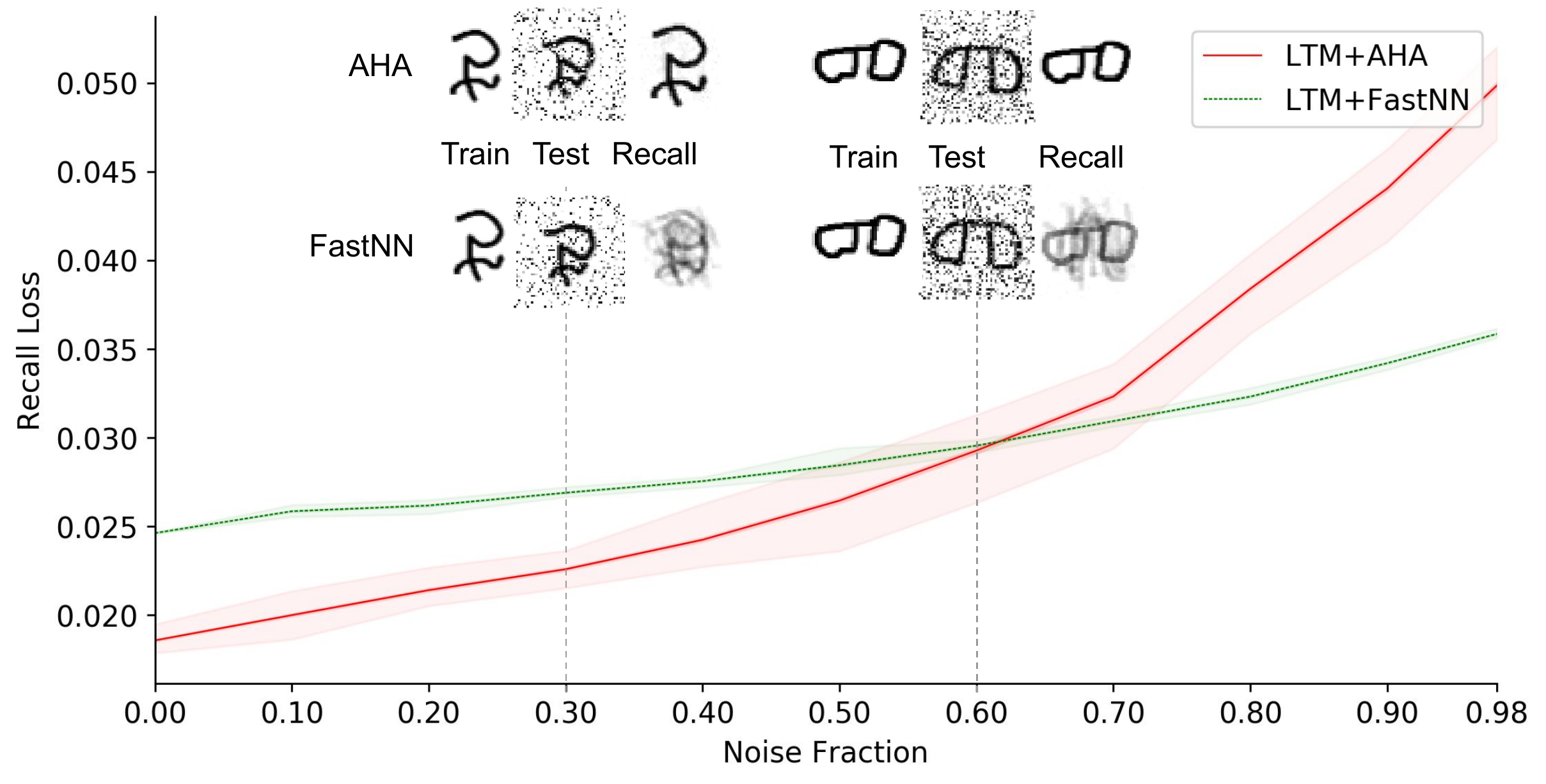}}

     \subfloat[\OneshotInst\ with occlusion]{\includegraphics[width=0.49\textwidth]{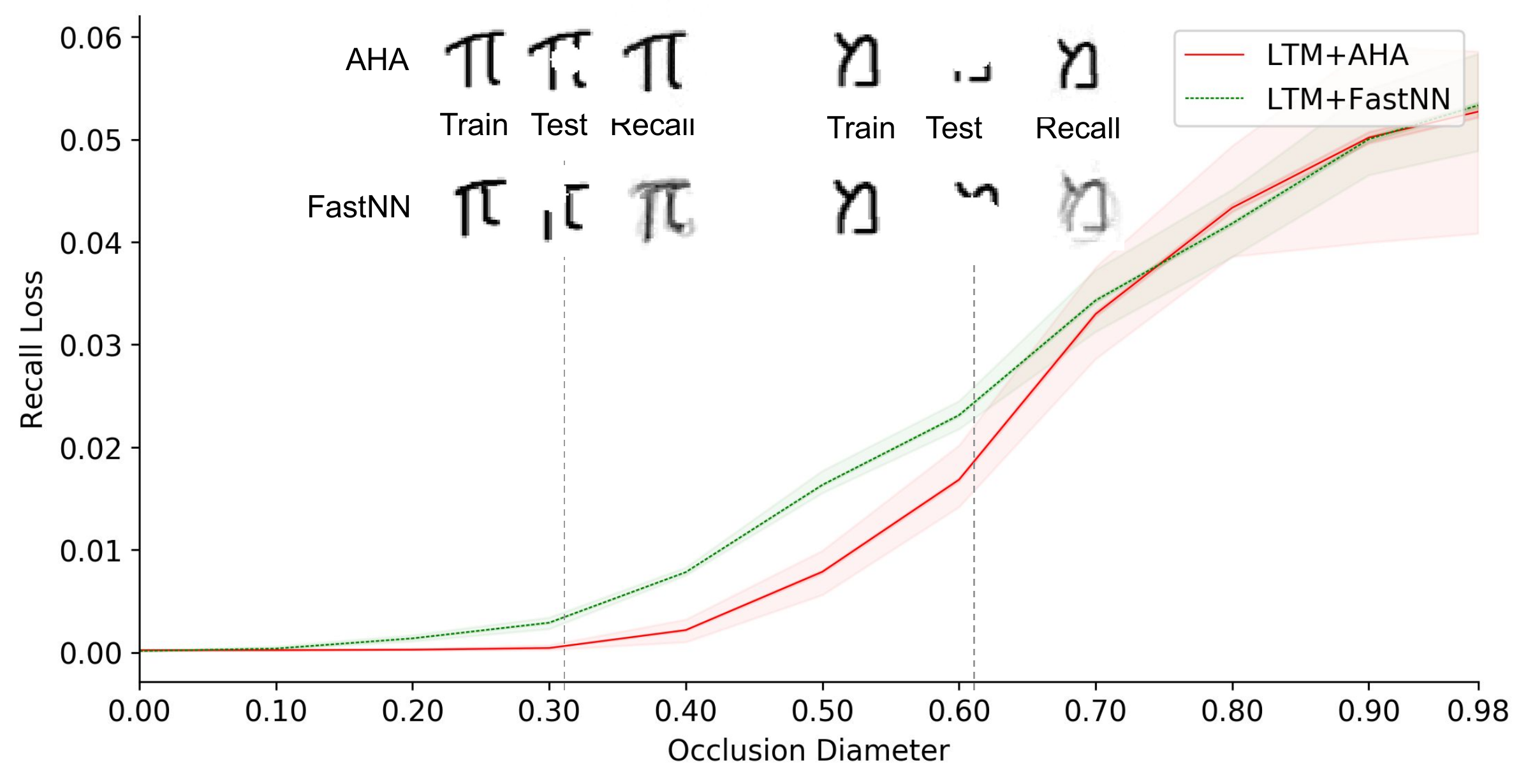}}
     \hfil
     \subfloat[\OneshotInst\ with noise]{\includegraphics[width=0.49\textwidth]{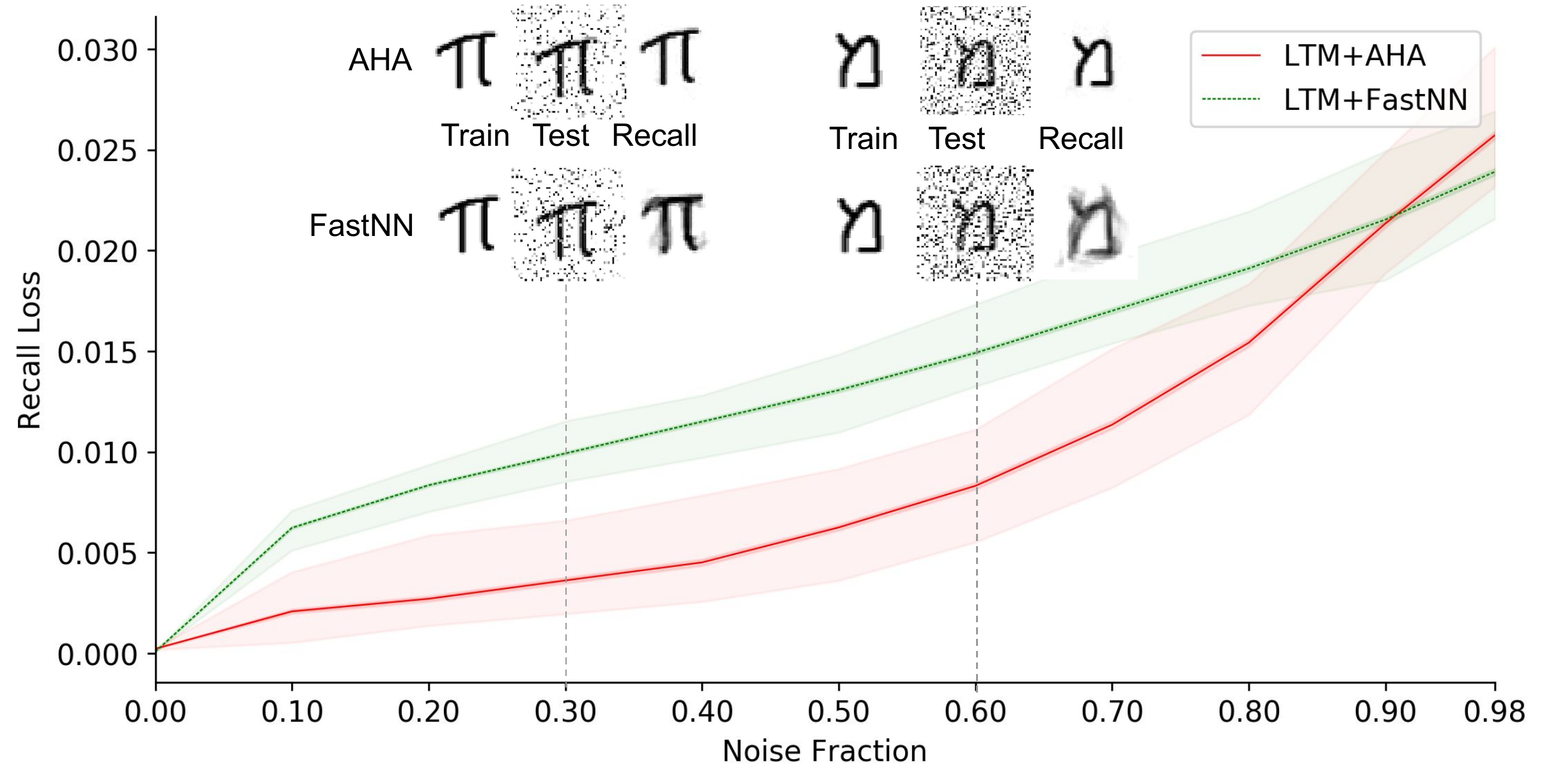}}

     \caption{\textbf{Recall-loss vs occlusion/noise.} AHA yields more specific and crisp recall images given moderate input corruption. With substantial corruption, AHA sometimes retrieves an accurate copy of the wrong character resulting in a higher loss. \FNN\ provides blurry or nonspecific recall in many conditions. Units and plot characteristics are the same as for Figure~\ref{fig:accuracy_plots}.}
     \label{fig:replay_loss}
\end{figure*}

Recall-loss is shown in Figure~\ref{fig:replay_loss}.
In the \oneshotGen\ experiment, AHA demonstrates better performance than \FNN\ under moderate occlusion and noise. At higher levels of corruption, AHA may retrieve a high quality image of the wrong character, resulting in a higher loss than lower-quality images retrieved by \FNN.
In the \oneshotInst\ experiment, this character confusion is less likely to occur and AHA is superior or equal to \FNN\ under all meaningful levels of image corruption.
\FNN\ is qualitatively better for \oneshotInst\ than for \oneshotGen, and almost as good as AHA, but recalls are typically an `average' version of the character, rather than a specific instances.


\section{Discussion}
\label{sec:discussion}

The results demonstrate that CLS is an effective approach to one-shot classification of both specific instances and categories.
At first sight, \oneshotInst\ is trivially solved by a nearest neighbour comparison, but the results show that this approach performs poorly given realistic levels of image corruption.
In addition, the CLS-style architecture of AHA has an advantage over the simpler \FNN. 
For accuracy, this is most noticeable on \oneshotGen\ and in the presence of occlusion, and it is significantly superior at recalling high quality images across the test conditions.
To the authors best knowledge, this is the first application of CLS to a dataset derived from real-world observations featuring non-synthetic variation.

AHA was comparable to state-of-the-art approaches on the standard Omniglot Benchmark - a subset of our extended test. The reported approaches are optimised for \oneshotGen\ without any image corruption suggesting that they are not suited for \oneshotInst.
Referring to Table~\ref{table:comparison}, BPL and RCN are significantly ahead of other methods, and similar to human performance. 
They have an advantage as they exploit domain specific priors as discussed in Section~\ref{sec:background_oneshot}.
The Simple Conv Net (CNN) represents a standard approach for deep learning. 
AHA is equally good despite being unsupervised (no external labels) and uses only local local credit assignment. Additionally, AHA demonstrates the broader range of capabilities discussed.

\PR\ performs classification significantly better than \PC.
It partially fulfils the role of completion, as it learns to reproduce the target.
\PC\ fulfils a vital role for additional completion and sharpening for crisp recall.
There is a small accuracy bias toward \PR\ due to the fact that \PR\ outputs a superposition of possible patterns, enhancing the chance of a correct match via MSE. In contrast, \PC\ is designed to retrieve a single, sharp complete sample and in doing so is unable to hedge its bets.


The boundary between class and exemplar is continuous, subjective and may depend on the task. 
For example, you could define the character itself as a class, and the corrupted samples as exemplars. Or a Labrador dog: the class could be the animal type (dog), or the breed (Labrador).
AHA demonstrates this flexibility to the task by accomplishing both \oneshotGen\ and \oneshotInst.
As per Section~\ref{sec:theory_of_operation_comp}, AHA learns a conjunction of primitives, and then generalises over variations in that combination. 

\section*{Acknowledgment}

Thanks to Elkhonon Goldberg for enriching discussions on the hippocampal region and to Rotem Aharon for insights and analysis of Hopfield networks.


\bibliographystyle{splncs04}
\bibliography{bibliography}

\end{document}